\documentclass[sigconf,nonacm]{acmart}
\usepackage{graphicx}
\usepackage{booktabs}

\usepackage{amsmath, amssymb}
\usepackage{multirow}
\usepackage{listings}
\usepackage{hyperref}
\usepackage{float}
\usepackage{graphicx}
\usepackage{tikz}
\usetikzlibrary{positioning}
\usepackage{stfloats}
\usetikzlibrary{positioning, arrows.meta, backgrounds}
\usepackage{caption}
\AtBeginDocument{%
  }

\renewcommand\footnotetextcopyrightpermission[1]{}
\setcopyright{acmlicensed}
\copyrightyear{2025}
\acmYear{2025}

\begin{document}

\title{CARE: Counselor-Aligned Response Engine for Online  Mental-Health Support}

\author{Hagai Astrin}
\affiliation{
   \institution{Ben-Gurion University}
   \country{}
}
\email{astrin@post.bgu.ac.il}

\author{Ayal Swaid}
\affiliation{
\institution{Ben-Gurion University}
   \country{}
}
\email{swaida@post.bgu.ac.il}

\author{Avi Segal}
\affiliation{
\institution{Ben-Gurion University}
   \country{}
}
\email{avise@post.bgu.ac.il}

\author{Kobi Gal}
\affiliation{
  \institution{Ben-Gurion University}
  \country{}
}
\affiliation{
  \institution{School of Informatics, University of Edinburgh}
  \city{Edinburgh}
  \country{UK}
}
\email{kobig@bgu.ac.il}

\renewcommand{\shortauthors}{Astrin et al.}

\begin{abstract}

Mental health challenges are increasing worldwide, leading to a growing demand for emotional support services. Providers of online mental health support frequently experience counselor overload, which can result in delayed responses during critical situations such as suicidal ideation, where timely, stabilizing contact is essential for the help-seeker’s safety. 
While large language models (LLMs) have shown strong generative capabilities, their use in low resource languages - especially in sensitive domains like mental health - remains largely underexplored. 
Moreover, existing LLM-based agents in this domain are typically not trained on large-scale, real-world datasets and therefore struggle to replicate the supportive language and intervention strategies practiced by established organizations. 
To address this gap, we propose CARE (Counselor-Aligned Response Engine for Mental-Health
Support), a GenAI-based framework that assists counselors by generating real-time, psychologically aligned response recommendations. 
CARE is based on fine-tunning open-source large language models separately for Hebrew and Arabic using curated subsets of real-world crisis conversations which reflect high-quality counseling practices. The training data is drawn from sessions rated as highly effective by professional counselors, enabling the models to capture interaction patterns associated with successful de-escalation and supportive interventions.
Our approach trains the models on complete conversation histories, allowing them to capture the evolving emotional context and the dynamic structure of counselor–help-seeker dialogue. 
In experimental settings, CARE demonstrates stronger semantic and strategic alignment with gold-standard counselor responses compared to vanilla, non-specialized studied LLMs.
These findings suggest that domain-specific fine-tuning on expert-validated data has the potential to support counselor workflows and contribute to improved care quality in low-resource language contexts.
\end{abstract}

\ccsdesc[500]{Computing methodologies~Natural language generation}

\keywords{Mental health, Large Language Models}


\maketitle

\section{Introduction}
Online mental health support has become an essential lifeline for people experiencing acute  distress~\cite{zalsman2021suicide}. 
In particular, in suicide prevention settings,   the right words, delivered at the right time, can literally save lives ~\cite{gutierrez2024examining,bhatt2024digital}. 
Online crisis hotlines and text-based services are experiencing a significant, sustained increase in usage globally, driven by heightened mental health awareness, the lasting impacts of the COVID-19 pandemic, and the normalization of digital support~\cite{zalsman2021suicide,daniels2025calls}.

Counselors are volunteers who receive an extensive training program and are supervised by mental health clinicians. 
These counselors are not just compassionate listeners - they are meticulously  trained to use  a psychologically informed conversational style to guide the discussion, which we refer to as  \emph{supportive language}~\cite{zalsman2021suicide}. 
This style blends empathy and  emotional validation  with crisis intervention principles in psychology research. 


 The significant increase in traffic on crisis hotlines   means that counselors are often required to conduct multiple concurrent conversations, each characterized by a distinct emotional trajectory, risk profile, and cultural-linguistic context~\cite{chiang2024using}.
Such high-pressure environments impose a heavy cognitive burden on counselors, increasing the risk of errors and response delays, which in turn can contribute to disengagement and dropout among help-seekers.

Recent advances in large language models (LLMs) offer new opportunities for adaptive user support in dialogue-based systems for mental health~\cite{hadarshoval2024assessing}. Nonetheless,  while general-purpose LLMs can generate fluent and empathetic text, they tend to produce generic reassurance or directive advice that is misaligned with the strategies used  by trained counselors using supportive language~\cite{Loh2023HarnessingLL,Iftikhar_Xiao_Ransom_Huang_Suresh_2025}.

For instance, a support-aligned response  (e.g., \textit{“I can't imagine the pain you are experiencing, I'm listening to whatever you want to say”}) acknowledges the speaker’s distress and communicates presence, while a superficially reassuring alternative generated by a baseline LLM  (e.g.,  \textit{“Don’t worry, it will be fine.”}) risks invalidating or minimizing their experience.

This work addresses  this gap by developing  a Counselor-Aligned Response Engine for Mental-Health Support (CARE). 
CARE is not intended to replace human judgment, but to serve as a  personalized decision-support tool  for counselors that generates real-time response suggestions grounded in professional counseling practice. The system accounts for the longitudinal interaction history between counselor and help-seeker, allowing it to adapt its recommendations to the evolving emotional and strategic context of the conversation.

CARE is trained using a large corpus  
of anonymized real-world crisis conversations from \emph{Sahar}, a national suicide-prevention chatline serving Arabic and Hebrew-speaking communities.\footnote{\url{https://sahar.org.il/about-us/}}  We fine-tune open-source LLMs on full conversation histories drawn from sessions rated as highly effective by the counselors. By supervising the model on complete dialogue trajectories, CARE implicitly learns how counselors use supportive language strategies over time, enabling it  to move beyond surface-level behaviour.


To evaluate CARE from a user modeling perspective, we compare its generated recommendations to gold-standard counselor responses from the dataset for the same conversational trajectory. In addition to standard semantic similarity metrics from the literature, we introduce the \emph{Support Intent Match (SIM)}, a novel measure that captures alignment between the intervention strategy expressed by CARE and that used by human counselors. This metric operationalizes whether the system has learned to adapt its responses to the inferred needs of the help-seeker at each point in the interaction.

To establish a robust performance benchmark, we compare CARE against unadapted, general-purpose baseline models. Our experiments involve three state-of-the-art open-source LLM implementations: Gemma-3-12B-it, Llama-3.1-8B-it and Qwen3-14B. We evaluate the framework across two distinct language-specific datasets, comprising professional counseling dialogues in    Arabic and Hebrew, to ensure that the adaptation process is effective across diverse morphological and cultural settings.

Our results demonstrate that CARE achieves significantly stronger semantic, stylistic, and strategic alignment with professional counselor behavior than unadapted baseline models, despite relying solely on implicit supervision from dialogue histories. These findings suggest that full-history fine-tuning enables LLMs to internalize user-adaptive interaction patterns without explicit strategy labels. 

More broadly, our work contributes to UMAP research by showing how longitudinal user modeling and domain-specific adaptation may support effective personalization in high-stakes conversational systems, particularly in low-resource language settings.


\section{Related Work}
Our work relates to several research areas in AI and mental health support: 
(1) LLMs for Mental Health Counseling
(2) Modeling Approaches for Dialogue Systems
(3) Linguistic Adaptation for Mental Health in Hebrew and Arabic.
We elaborate on each of these areas below.

\subsection{LLMs for Mental Health Counseling}
Recent advances have explored large language models for direct engagement in mental health contexts, with a focus on reliability, empathy, and safety. 
Lan et al.~\cite{lan2024reliable} developed a depression-diagnosis-oriented conversational system that balances clinical reliability with emotional warmth through a dual-ontology framework. Their approach integrates task-oriented symptom identification based on DSM-5 criteria with empathetic strategies rooted in Helping Skills Theory to improve both diagnostic precision and the therapeutic alliance.
Kampman et al.~\cite{kampman2022multiagent} proposed a multi-agent dual-dialogue framework in which AI agents assist mental health care providers by supplying relevant knowledge and conversation suggestions in real time. Their approach focuses on augmenting, rather than replacing, human counselors.

Welivita and Pu~\cite{welivita2024heal} contributed the HEAL knowledge graph, encoding domain knowledge from distress management conversations to enable structured reasoning over conversational content. 
Xu et al.~\cite{xu2025evaluation} conducted a large-scale benchmark of 15 state-of-the-art LLMs, including GPT-4.1, DeepSeek-R1, Llama4, and Qwen QwQ, on tasks of mental health knowledge testing and illness diagnosis in the Chinese context. Their findings highlight both the promises and limitations of current LLMs in sensitive mental health applications, underscoring the need for domain-adapted models.
While these works address important safety, augmentation, and knowledge structuring needs, none have investigated fine-tuning LLMs on large-scale Hebrew or Arabic suicide-prevention chat data, nor have they focused on aligning models with the support strategies and language of a specific organization, which is the focus of our work.

\subsection{Modeling Approaches for Dialogue Systems}
Researchers have proposed multiple approaches to model conversations in dialogue systems. \citet{zhang2024advancingconversationalpsychotherapyintegrating} proposed a psychotherapy-oriented LLM framework that incorporates a dual-memory architecture to balance short- and long-term context, while embedding domain expertise and privacy safeguards to improve reliability and therapeutic safety.
Wang et al.\cite{wang2025recursivelysummarizingenableslongterm} introduced a multi-session paradigm for automated counseling, utilizing a top-down generation approach to maintain consistency and capture therapeutic progression across multiple sessions.
By modeling these extended interactions, the system aims to emulate how human counselors dynamically recalibrate goals based on a client's evolving mental state over time.
Wang et al.~\cite{wang2025recursivelysummarizingenableslongterm} proposed a recursive summarization mechanism that incrementally condenses dialogue history into long-term memory. 
This approach improves consistency in extended conversations and complements both long-context and retrieval-enhanced models, offering a lightweight strategy for maintaining dialogue coherence over time.

Notably, these works primarily address dialogue settings that span multiple sessions and rely on explicit memory, retrieval, or continual adaptation mechanisms. In contrast, our work focuses on anonymous, single-session conversations between a help-seeker and a counselor. We study the efficacy of full-history supervised fine-tuning on a large, domain-specific corpus, investigating how modeling the entire conversational trajectory within a single session enables an LLM to internalize professional counseling patterns without the need for external retrieval, persistent memory modules, or online adaptation.


\subsection{Linguistic Adaptation for Mental Health in Hebrew and Arabic}
The application of Natural Language Processing (NLP) to mental health in morphologically rich languages like  Arabic or Hebrew presents unique structural and cultural challenges. Most foundational research in this domain has historically focused on English-centric datasets such as Reddit \cite{low2020natural}. We mention advancement for each language in turn.  {HeBERT} and {HebEMO} \cite{chriqui2021hebert}   provided the first robust transformer-based tools for sentiment and emotion recognition in modern Hebrew. 
\citet{izmaylov24} introduced SR-BERT (Suicide Risk-BERT), a hierarchical transformer model that integrates psychological theory with dialogue structure to predict suicide risk in real-time crisis chats. Similarly, \citet{elyoseph2025applying} explored the use of LLMs to evaluate Hebrew media adherence to suicide reporting guidelines, highlighting the growing capability of generative models in the Hebrew suicide prevention ecosystem. Despite these advancements, most existing Hebrew models are discriminative - focused on classification and risk detection, or rely on zero-shot prompting of commercial LLMs. Our work addresses this gap by providing the first generative, domain-specific fine-tuning of a large-scale model (Gemma-3-12B-it) on professional Hebrew counseling dialogues, focusing on the production of supportive language rather than just risk classification.

Research in Arabic Natural Language Processing (NLP) for mental health faces two central challenges: the language’s morphological richness and its extensive dialectal diversity. While early work was limited by the scarcity of Dialectal Arabic (DA) data, recent research increasingly incorporates dialect-aware models to better handle this linguistic complexity. Initial efforts include classical machine learning approaches for depression detection on DA tweets \cite{ALMOUZINI2019257}, followed by RNN-based models \cite{Alabdulkreem03072021} and attention-based BiLSTM architectures \cite{cmc.2022.022609} for depression detection.

With the advent of transformers, studies have demonstrated the robustness and performance of models such as AraBERT and MARBERT~\cite{10570345,s22030846} in detecting depression and other mental health conditions. 
Further work has focused on suicide ideation prediction, highlighting the effectiveness of encoder-only models such as MARBERT and the Universal Sentence Encoder (USE) on social media data~\cite{Baghdadi2022SuicideArabicTweets, ALATAWI2024143} .
Beyond classification, the MentalQA dataset \cite{10600466_a} provides question--answer pairs between help-seekers and professional doctors, with annotations for question types and response strategies. It has been used for generative question answering in mental health, including a multi-agent framework with retrieval-augmented generation (RAG) by Sabty et al.~\cite{sabty2025fahmni}, and a culturally aware RAG pipeline by AbdelAziz et al.~\cite{abdelaziz2025arabic}.

\begin{figure*}[t]
  \centering
  \includegraphics[width=\textwidth]{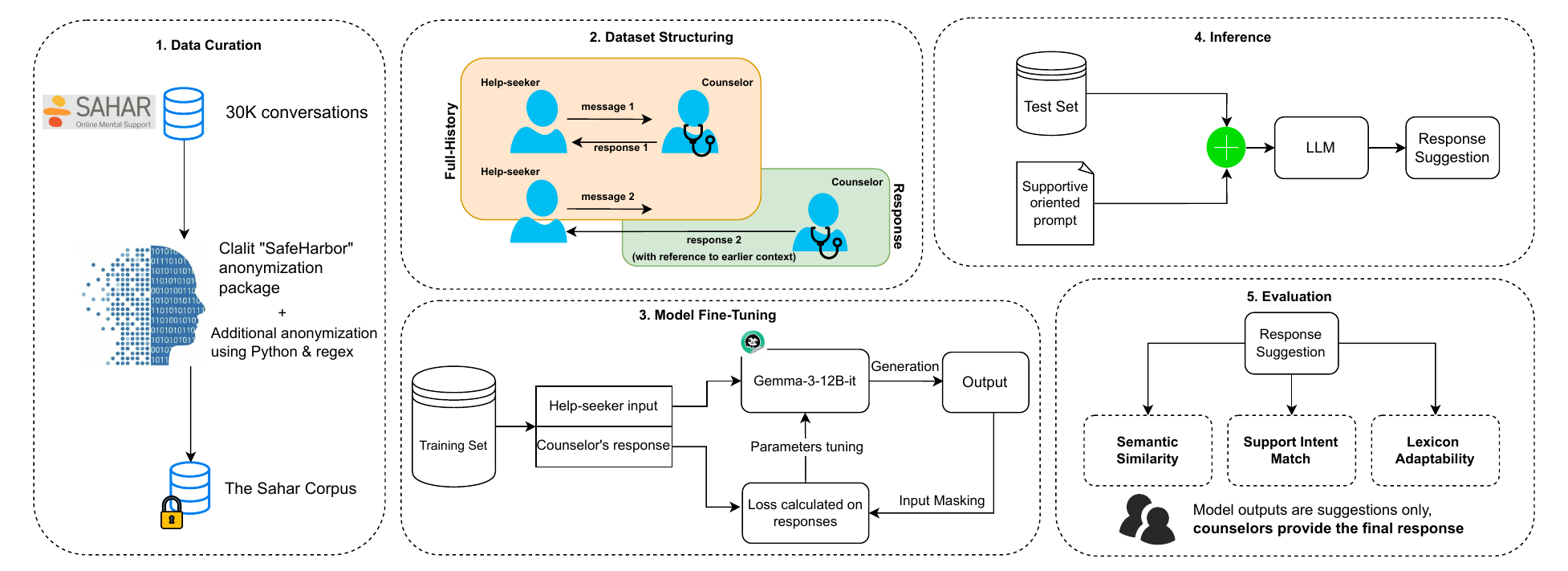}
  \caption{The CARE Framework}
  \label{fig:full_methodology}
\end{figure*}

\section{The Sahar Domain}
Sahar\footnote{https://sahar.org.il/en/} (acronym for “Aid and Attention Online”)
is a leading online mental health chat line serving Arabic- and Hebrew-speaking communities, with a focus on suicide prevention and emotional distress relief. Alleviating the emotional distress of help-seekers is a crucial step in the process of suicide prevention~\cite{overholser1997,suris1996}. 
The organization handles more than 50,000 chat sessions a year,
and these numbers are continuing to increase.

Sahar’s counselors are volunteers over 24 years old that completed a special training program by licensed clinicians. 
They are trained in specialized crisis intervention protocols that prioritize active listening and adaptive communication strategies tailored to the evolving emotional trajectory of the help-seeker. 
During each shift, therapists are also on duty to monitor conversations and provide professional guidance to counselors when needed. 
At the end of each session, counselors conduct a clinical debrief by providing a summary of the dialogue and completing a structured survey. This assessment identifies the primary thematic content and categorizes the help-seeker’s suicide risk into a three-tier hierarchical system:
  {General Suicide Risk (GSR):} Assigned when a help-seeker exhibits clinical signs of suicidal ideation, such as expressions of hopelessness or thoughts of self-harm.
 {Immediate Suicide Risk (IMSR):} A critical subset of GSR ($IMSR \subset GSR$), identifying cases of imminent danger where the individual may act on suicidal intentions at any moment.
 {No Risk:} Assigned when no indicators of suicidal intent are present.

Sahar’s counselors are trained in specialized crisis intervention protocols that prioritize active listening and adaptive communication strategies. Central to this approach is the use of specific support strategies, which are employed to guide the help-seeker through different stages of the de-escalation process:
\textbf{Reflection:} Mirroring the help-seeker's expressed emotions or thoughts to demonstrate understanding and validate their experience (e.g., ``It sounds like you are feeling very overwhelmed right now.'').
\textbf{Prompting:} Using open-ended questions or gentle nudges to encourage the help-seeker to share more details about their situation or feelings (e.g., ``Can you tell me more about what happened today?'').
\textbf{Suggestion:} Providing gentle guidance or proposing positive coping mechanisms and next steps, once the help-seeker feels heard and validated (e.g., ``Perhaps we could think of one small thing that might make you feel safer tonight?'').

Our research utilizes a comprehensive corpus of crisis intervention dialogues provided by the Sahar organization, 
spanning two languages: Hebrew and Arabic.
This dual-language dataset allows us to develop and evaluate the CARE framework across diverse linguistic and cultural contexts, addressing the critical need for localized mental health support.

\subsection{Arabic/Hebrew Crisis Dataset}
The Hebrew dataset consists of 30,232 anonymous chat sessions collected between 2019 and 2023. This corpus comprises 1,486,694 messages, averaging 49.17 messages per interaction. 
In terms of risk distribution, 5,163 sessions ($\approx17\%$) are labeled as GSR.

The Arabic dataset consist of 2,376 anonymous chat sessions collected between 2021 and 2025.
This corpus comprises 132,167 messages, averaging 55.62 messages per interaction.
In terms of risk distribution, 227 sessions ($\approx9.5\%$) are labeled as GSR. \\

To evaluate the clinical impact of these sessions, the organization utilizes the \emph{Volunteer Effectiveness Dimension (VED)}. The VED is a clinical label filled out by the counselor conducting the chat at the conclusion of each session, and it reflects their assessment of the quality and effectiveness of the support provided. Ratings are given on a 1-5 scale, where 1 indicates that the counselor was not able to assist the help-seeker, and 5 represents optimal de-escalation and assistance.

\section{The CARE System}
The {Counselor-Aligned Response Engine (CARE)} system is designed to produce psychologically aligned, context-aware response recommendations in online crisis hotlines. The system architecture is illustrated in Figure~\ref{fig:full_methodology}. 

As seen in the figure, our methodology follows a five-stage pipeline: (1) Data Curation: filtering the Sahar corpus for interventions with documented clinical success; (2) Dataset Structuring: Transforming raw chats into a sequential format where each intervention is conditioned on the full preceding dialogue history; (3) Model Fine-Tuning: adapting open-source LLMs using a full-history masked loss objective to capture longitudinal dialogue dynamics; (4) Inference: Generating response suggestions by prompting the fine-tuned model with a supportive system prompt and the complete test-set history; and (5) Evaluation: Benchmarking suggestions against expert responses using semantic, lexical, and clinical metrics, including the new Support Intent Match (SIM).

\subsection{Data Selection and Curation}
To ensure the CARE system internalizes the most effective recommendation strategies, 
we restricted our analysis to a curated subset of 1,000 sessions that achieved a VED score of 4 or 5. This high-effectiveness corpus was partitioned into 800 sessions for training and 200 sessions for evaluation in each of the two languages. 

The data underwent a two-tier anonymization process to protect help-seeker privacy (Figure \ref{fig:full_methodology}, Step 1: Data Curation). In Hebrew, we applied \texttt{HebSafeHarbor} package\footnote{\url{http://github.com/ChenMordehai/HebSafeHarbor_Clalit_Validation_Improvment}}, a specialized de-identification package developed by Clalit Health Services - Israel's largest healthcare provider, specifically for the automated removal of Protected Health Information (PHI) from Hebrew clinical and conversational text.  
In Arabic, we used \texttt{CAMeLBERT MSA NER} model\footnote{\url{https://huggingface.co/CAMeL-Lab/bert-base-arabic-camelbert-msa-ner}}, a Named Entity Recognition (NER) model that was built by fine-tuning the CAMeLBERT Modern Standard Arabic (MSA) model\footnote{\url{https://huggingface.co/CAMeL-Lab/bert-base-arabic-camelbert-msa}}.

To preserve the longitudinal context of the crisis interactions, the sessions were structured into training samples containing the full dialogue history up to each counselor intervention (Figure~\ref{fig:full_methodology}, Step 2: Dataset Structuring). This structuring method expanded the 800 training sessions into 12,100 unique samples for the Hebrew corpus and 9,970 for the Arabic corpus. Similarly, the 200 evaluation sessions were transformed into 3,097 Hebrew and 2,997 Arabic test samples. By utilizing these full-history trajectories, the model is trained to integrate multi-turn emotional cues into its response generation process, replicating the holistic assessment performed by Sahar's counselors.

For both datasets, we used a secondary stage of regex-based filtering to remove any remaining identifiers not captured by the initial package, such as specific local phone formats and service-specific metadata.

\subsection{Model Training}
We fine-tuned the Gemma-3-12B-it model using Supervised Fine-Tuning (SFT), which enabled training on full conversation histories using a single Nvidia RTX4090 GPU (24GB vRAM). To optimize the training pipeline, we utilized the \texttt{Unsloth}\footnote{\url{https://unsloth.ai/}} framework alongside LoRA\cite{hu2021lora}, significantly reducing memory overhead while preserving the model's complex reasoning capabilities.

\paragraph{Training Objective and Masked Loss:} 
Training was conducted on the 12,100 Hebrew utterances and 9,970 Arabic utterances, derived from chat sessions with high clinical effectiveness. We employed a masked loss approach to focus the learning signal strictly on the counselor’s interventions (Figure \ref{fig:full_methodology}, Step 3: Model Fine-Tuning). In this configuration, the model is provided with the full dialogue history - treated as a fixed prefix, alongside a domain-specific system prompt designed to ensure empathetic and safe phrasing in both Hebrew and Arabic.

The model is only penalized for its predictions of the counselor's response tokens. This prevents the model from wasting its learning capacity on predicting the help-seeker's messages or generic conversational fillers. Instead, this approach forces the model to prioritize the specific supportive language patterns and de-escalation strategies characteristic of professional counselors, ensuring that the generated suggestions are grounded in clinical expertise rather than general-purpose dialogue.

\paragraph{Hyperparameter Configuration:} 
We fine-tuned the model using Low-Rank Adaptation (LoRA)~\cite{hu2021lora}, setting both the rank ($r$) and scaling factor ($\alpha$) to 8. These values align with the optimized defaults recommended by the \texttt{Unsloth} library for stable and memory-efficient domain adaptation. Setting $\alpha=r$ ensures a normalized scaling of the low-rank updates, which stabilizes the training process and prevents the model from diverging from its base linguistic capabilities~\cite{yang2022featurelearning}.

\subsection{Inference and Prompting}
At inference time, the model is provided with the full dialogue history and a language-specific system prompt designed to ensure empathetic and safe phrasing (Figure \ref{fig:full_methodology}, Step 4: Inference). These prompts, developed for both Hebrew and Arabic, serve as core behavioral constraints, codifying Sahar’s organizational values to maintain a supportive and non-judgmental environment. 
Below is the English reference version of the prompt used to guide model behavior across both language implementations.

\begin{quote}
“You are an empathetic and compassionate emotional support assistant, providing emotional assistance through text conversations to people seeking help. Your role is to actively listen, validate their feelings, and offer emotional support. Gently encourage, ask open-ended questions, and guide help-seekers toward positive coping strategies. Avoid making clinical diagnoses or giving medical treatment recommendations. If a help-seeker states that they are in immediate distress or mentions self-harm, gently suggest contacting a mental health professional or reaching out to emergency services. Always maintain a safe and non-judgmental space where help-seekers can share openly.”
\end{quote}

\section{Evaluation}
Evaluation compares model-generated responses with gold-standard replies provided by human counselors for the same conversational contexts (Figure~\ref{fig:full_methodology}, Step 5: Evaluation).
Evaluation is conducted on the held-out test-set: 3,097 and 2,997 utterances in Hebrew and Arabic respectively. 
We note that state-of-the-art proprietary LLMs (e.g., GPT-4o, Claude) were not used, as organizational privacy constraints and the sensitivity of crisis data require all evaluation to be conducted on locally hosted models.
The following metrics were used for evaluation:

\subsection{Semantic and Lexical Similarity}
Alignment between model-generated responses and gold-standard counselor replies was assessed using:
\begin{itemize}
    \item \textbf{BERTScore}~\cite{zhang2020bertscoreevaluatingtextgeneration}: This metric measures semantic similarity using contextual embeddings, capturing whether the underlying meaning of the response is preserved even when different wording is used. We report the average F1 BERTScore.
    
    \item \textbf{ROUGE (1, 2, L)}~\cite{lin2004rouge}: These metrics quantify recall by measuring overlap in n-grams and structural sequences between generated and reference responses. In particular, ROUGE-L captures the longest common subsequence, reflecting whether the model preserves the structural flow and coherence of a professional crisis intervention.
    
    \item \textbf{BLEU}~\cite{papineni2002bleu}: While ROUGE emphasizes recall, BLEU measures precision by penalizing the inclusion of irrelevant or redundant tokens not present in the counselor’s response. This serves as a proxy for fluency, conciseness, and lexical precision relative to the human expert.
\end{itemize}

\subsection{Support Intent Match (SIM)}
Sahar’s support methodology is grounded in active listening and context-sensitive emotional support. We propose the SIM metric to quantify alignment between the support strategy expressed by the model and that of the gold-standard reference. Support strategies are categorized into predefined classes defined by the Sahar organization. These include \textit{Reflection}, \textit{Suggestion}, \textit{Prompting}, and \textit{Neutral} capturing the clinical intent behind each counselor utterance.

To perform this evaluation, we developed a Strategy Classifier for identifying the support strategy expressed in a given counselor message. This classifier is used to analyze and compare the strategies of model-generated responses and gold-standard references.

The Strategy Classifier was built using a hierarchical labelling and training pipeline. We first constructed a gold-standard set of 150 chat sessions manually annotated by a certified psychologist, yielding 5,038 counselor responses with strategy labels. A vanilla \textit{Gemma-3-12B-it} model was then evaluated in a zero-shot setting on this set~\cite{elyoseph2025applying, hua2024applyingevaluatinglargelanguage}, achieving F1-scores of 72.6\% for \textit{Reflection}, 78.2\% for \textit{Prompting}, 49.1\% for \textit{Suggestion} and 70.4\% for \textit{Netural}. Based on this performance, the LLM was used to generate silver-standard labels for the full Sahar Hebrew and Arabic datasets. Using these labels, we trained two language-specific classifiers in a supervised manner, where the input $x$ is a counselor utterance and the target is one of four strategies: Reflection, Prompting, Suggestion or Neutral. We fine-tuned AlephBERT~\cite{seker2021alephbertahebrewlargepretrained} for Hebrew and AraBERTv2~\cite{antoun2020arabert} for Arabic
, with batch size 64, learning rate $3 \times 10^{-5}$, for 15 epochs, and early stopping based on validation F1-score.

The SIM metric was computed as the proportion of matching strategies (as identified by the Strategy Classifier) across the evaluation set:
\begin{equation}
\text{SIM} = \frac{1}{M} \sum_{j=1}^{M} \left[\hat{s}_j = s_j\right]
\end{equation}

\noindent where $\hat{s}_j$ denotes the support strategy predicted for the generated response, $s_j$ is the ground-truth strategy, and $M$ is the number of samples. For a given model, SIM reflects how well the model has implicitly learned to select appropriate counseling strategies based solely on conversational context.

For an unadapted baseline model, the metric captures the inherent strategic capability of a general-purpose LLM guided only by a supportive system prompt. In this case, SIM measures the model’s zero-shot ability to generate interventions aligned with professional counselor strategies, without explicit fine-tuning on the Sahar corpus.

\subsection{Supportive Language Perplexity (PPL)}
To assess stylistic alignment with Sahar’s professional support language, we compute perplexity~\cite{bengio2003neural} using an independent language model trained exclusively on counselor utterances from our dataset. Concretely, a GPT-2 language model was trained on counselor responses extracted from the Sahar corpus using standard next-token prediction with cross-entropy loss, following established practice in language modeling for stylistic evaluation\footnote{\url{https://huggingface.co/docs/transformers/perplexity}}.

The perplexity assigned by the trained GPT-2 model to an utterance is defined as:
\begin{equation}
\text{PPL} = \exp\left(-\frac{1}{N} \sum_{i=1}^{N} \log p(w_i) \right)
\end{equation}

\noindent where $w_i$ denotes the $i$th token and $N$ is the total number of tokens. Perplexity for a given condition is computed by averaging the scores across all utterances generated under that condition. Lower perplexity values indicate stronger conformity to the stylistic and linguistic patterns characteristic of professional crisis counselors.

Our evaluation applies these metrics to compare CARE with multiple unadapted baseline models, including Gemma-3-12B-it, Qwen3-14B, and Llama-3.1-8B-it, all operating under the same supportive system prompt. This design isolates the effect of domain-specific fine-tuning and quantifies its contribution in moving from generic AI responses toward clinically grounded, empathetic support.

\section{Results}
The CARE model demonstrates a significant performance advantage over the unadapted Gemma-3-12B-it, Qwen3-14B, and Llama-3.1-8B-it baselines across all clinical and linguistic dimensions. While the baseline models, guided by supportive system prompts, produce grammatically correct Hebrew and Arabic responses, they consistently fail to replicate the specific strategic and stylistic nuances found in professional Sahar interventions.

\subsection{Support Intent Match}
\begin{figure}[h]
    \centering
    \includegraphics[width=\linewidth]{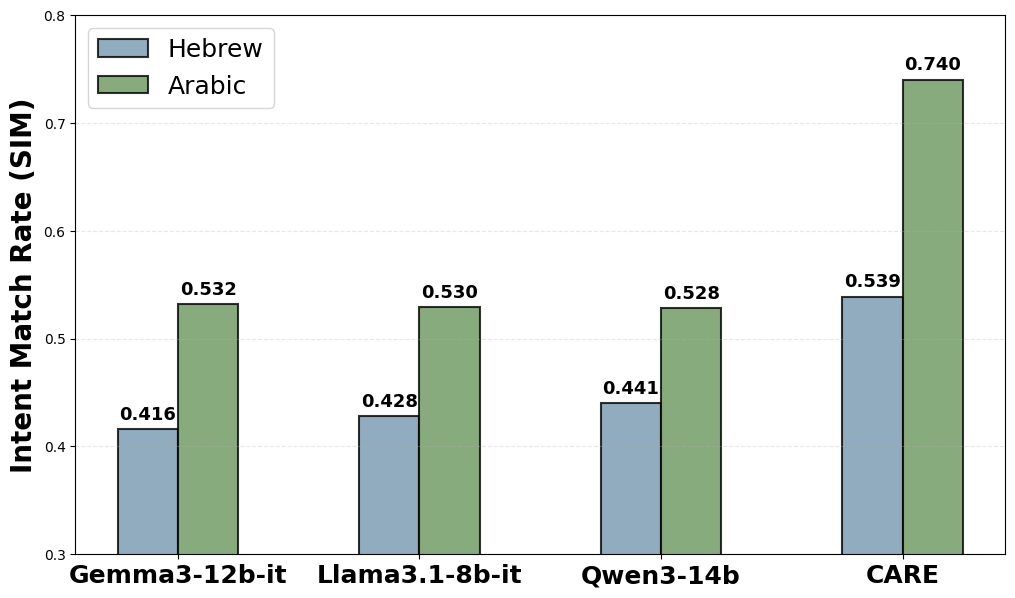}
    \caption{Comparison of Support Intent Match (SIM) scores between baseline models and CARE. The prediction task evaluates the model's ability to correctly identify the specific counseling strategy utilized in gold-standard responses, where a higher score indicates greater strategic alignment with professional counselor behaviors.}
    \label{fig:SIM}
\end{figure}

Figure~\ref{fig:SIM} presents the SIM metric, evaluating the agreement between a model’s generated strategy and the strategy of a gold-standard counselor's response as identified by the Strategy Classifier. As shown in the figure, CARE achieves a match score of 0.53 in Hebrew and 0.74 in Arabic, significantly outperforming the baselines scores. It is important to note that since CARE was not explicitly trained on intent labels, this result indicates that the model implicitly learned to select appropriate counselling strategies purely through exposure to the longitudinal flow of dialogue in the training set.

\subsection{Semantic and Lexical Similarity}

\begin{figure}[h]
    \centering
    \includegraphics[width=\linewidth]{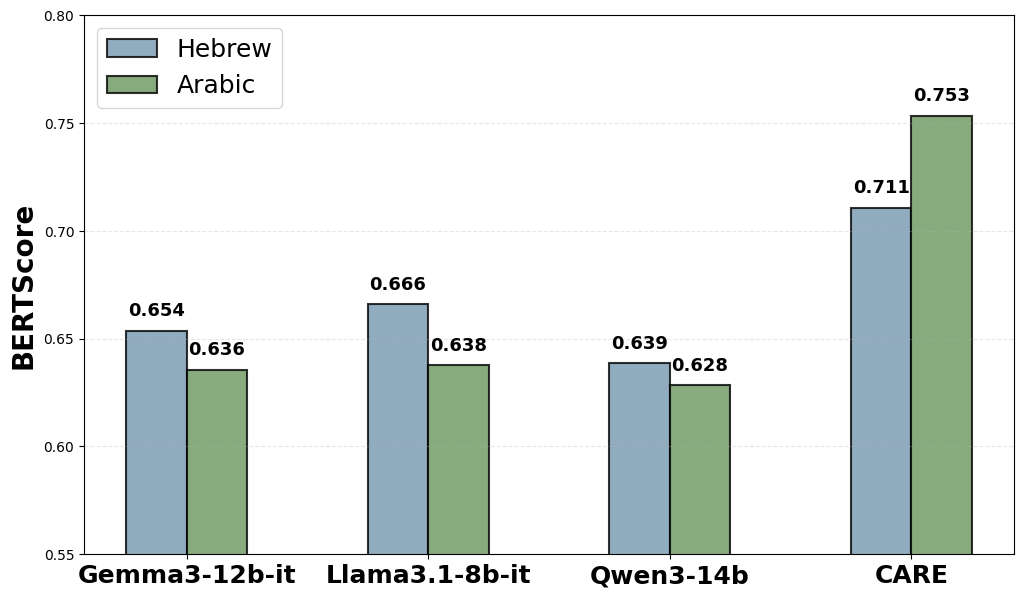}
    \caption{Comparison of BERTScore scores between the baseline models and CARE. The metric measures the semantic alignment between model-generated responses and gold-standard counselor interventions, utilizing contextual embeddings to capture shared meaning even when lexical choices differ.}
    \label{fig:BERTScore}
\end{figure}

As shown in Figure~\ref{fig:BERTScore}, CARE demonstrates a substantial improvement in both languages over the baseline models across all similarity dimensions. The BERTScore increased significantly, indicating that the fine-tuned model's responses are semantically closer to the expert counselor's responses.

The capacity of CARE to capture the specific content and structural sequences of professional counseling is further evidenced by the ROUGE scores. As shown in table~\ref{tab:linguistic-metrics}, the model achieved a ROUGE-1 score of 0.23 in Hebrew and 0.32 in Arabic, doubling the content-word recall of the baseline models in Hebrew and providing a nearly fivefold improvement in Arabic compared to the baseline average. The ROUGE-L scores, which rose to 0.21 in Hebrew and 0.31 in Arabic, further indicate that the fine-tuning process successfully internalized the longitudinal structural coherence and professional response architecture used by human experts, compared to the baseline approaches. Additionally, the Hebrew ROUGE-2 score exhibited a nearly ninefold increase - rising from 0.0123 to 0.1109, while the Arabic score reached 0.23, suggesting that CARE better internalizes the distinct bigram linguistic signature of the Sahar organization.

\begin{table*}[t]
\centering
\caption{Linguistic Overlap Metrics: Comparison of ROUGE and BLEU Scores}
\label{tab:linguistic-metrics}
\small
\begin{tabular}{lcccc|cccc}
\toprule
 & \multicolumn{4}{c}{\textbf{Hebrew}} & \multicolumn{4}{c}{\textbf{Arabic}} \\
\cmidrule(lr){2-5} \cmidrule(lr){6-9}
\textbf{Model} & \textbf{R-1} & \textbf{R-2} & \textbf{R-L} & \textbf{BLEU} & \textbf{R-1} & \textbf{R-2} & \textbf{R-L} & \textbf{BLEU} \\
\midrule
Gemma3-12b-it & 0.1383 & 0.0123 & 0.1049 & 1.1067 & 0.0745 & 0.0020 & 0.0641 & 0.7089 \\
Llama3.1-8b-it & 0.1503 & 0.0221 & 0.1263 & 2.0686 & 0.0638 & 0.0059 & 0.0574 & 0.8600 \\
Qwen3-14b & 0.1104 & 0.0095 & 0.0876 & 0.7771 & 0.0645 & 0.0016 & 0.0562 & 0.5703 \\
\midrule
\textbf{CARE} & \textbf{0.2346} & \textbf{0.1109} & \textbf{0.2167} & \textbf{6.0689} & \textbf{0.3272} & \textbf{0.2351} & \textbf{0.3116} & \textbf{20.4463} \\
\bottomrule
\end{tabular}
\end{table*}

The gains in precision and fluency are notably reflected in the BLEU score results in table~\ref{tab:linguistic-metrics}, which serve as a proxy for the lexical accuracy of the model's output \cite{papineni2002bleu}. here, the CARE model demonstrates a significant performance gain; the score reached 20.446 in Arabic, which is over 20 times higher than the baseline models. In Hebrew, the model achieved a score of 6.068, representing a result six times better than the baseline average. These results demonstrate that while unadapted models struggle with generating accurate and fluent professional responses in these languages, CARE’s outputs exhibit significantly higher precision alignment with the specific lexical and phrasal patterns of trained counselors. Collectively, these metrics validate that domain-specific adaptation effectively decreases the gap between generic AI dialogue and the authentic, clinically-aligned language required for professional crisis de-escalation.

\subsection{Supportive Language Usage: Perplexity}
We evaluate stylistic conformity using the perplexity (PPL) metric, computed with a GPT-2 language model trained on sampled counselor utterances from the Sahar corpus. Lower perplexity indicates a stronger resemblance to the tone and phrasing typical of counselors. 
As shown in Table~\ref{tab:style-metrics}, CARE demonstrates superior stylistic alignment across both languages compared to all unadapted baselines. In Hebrew, CARE achieved the lowest perplexity of 3.3383, successfully outperforming general-purpose models. The improvement in Arabic was even more pronounced; CARE reached a PPL of 2.6324, representing a nearly 60\% reduction in perplexity compared to the baseline models. 
These results confirm that the CARE model better internalizes the distinct, validating, and reflective tone characteristic of trained Sahar counselors. While the unadapted baselines often produce generic or prescriptive clinical advice, the low PPL scores for CARE propose that our domain-adapted model produces text that more closely mirrors the support organization style.

\begin{table}[h]
\centering
\caption{Supportive Language Style Metric: Comparative Perplexity}
\label{tab:style-metrics}
\begin{tabular}{lcc}
\toprule
& \multicolumn{2}{c}{\textbf{Supportive Language Perplexity ($\downarrow$)}} \\
\cmidrule(lr){2-3}
\textbf{Model} & \textbf{Hebrew} & \textbf{Arabic} \\
\midrule
Gemma3-12b-it  & 3.5246 & 7.2426 \\
Llama3.1-8b-it & 4.9406 & 6.8674 \\
Qwen3-14b      & 4.3312 & 6.7991 \\
\midrule
\textbf{CARE} & \textbf{3.3383} & \textbf{2.6324} \\
\bottomrule
\end{tabular}
\end{table}

\subsection{Statistical Significance Testing}
Statistical comparisons were conducted separately for the Hebrew and Arabic test sets. 
For each evaluation metric, we compared CARE against each baseline model (Gemma3-12B-it, Qwen3-14B, and Llama3.1-8B-it) using paired Wilcoxon signed-rank tests, which are appropriate for bounded and non-normally distributed evaluation measures. 
To control for multiple hypotheses, Holm correction was applied within each metric over the three CARE–baseline comparisons.

Across both languages, CARE differed significantly from all three baselines on all core similarity and overlap metrics, including ROUGE-1, ROUGE-2, ROUGE-L, BLEU, and BERTScore (Holm-corrected \(p \approx 0\) in all cases). 
The SIM metric also showed a highly significant advantage for CARE in both Hebrew and Arabic, with corrected p-values ranging from \(10^{-19}\) to \(10^{-23}\) for Hebrew and from \(10^{-69}\) to \(10^{-74}\) for Arabic.

For Supportive Language Perplexity, which measures alignment with clinically appropriate terminology, CARE was significantly different from Gemma3-12B-IT, Qwen3-14B, and Llama3.1-8B-IT in both languages (Holm-corrected \(p \approx 0\)), indicating a robust and consistent distinction in domain-specific language usage.

Taken together, the Wilcoxon analysis demonstrates that the performance differences between CARE and the LLM baselines are highly significant, stable across languages, and consistent across all evaluation dimensions, providing strong statistical evidence for the robustness of CARE’s advantage.

\subsection{Sensitivity Analysis}
\begin{figure}[h]
    \centering
    \includegraphics[width=\linewidth]{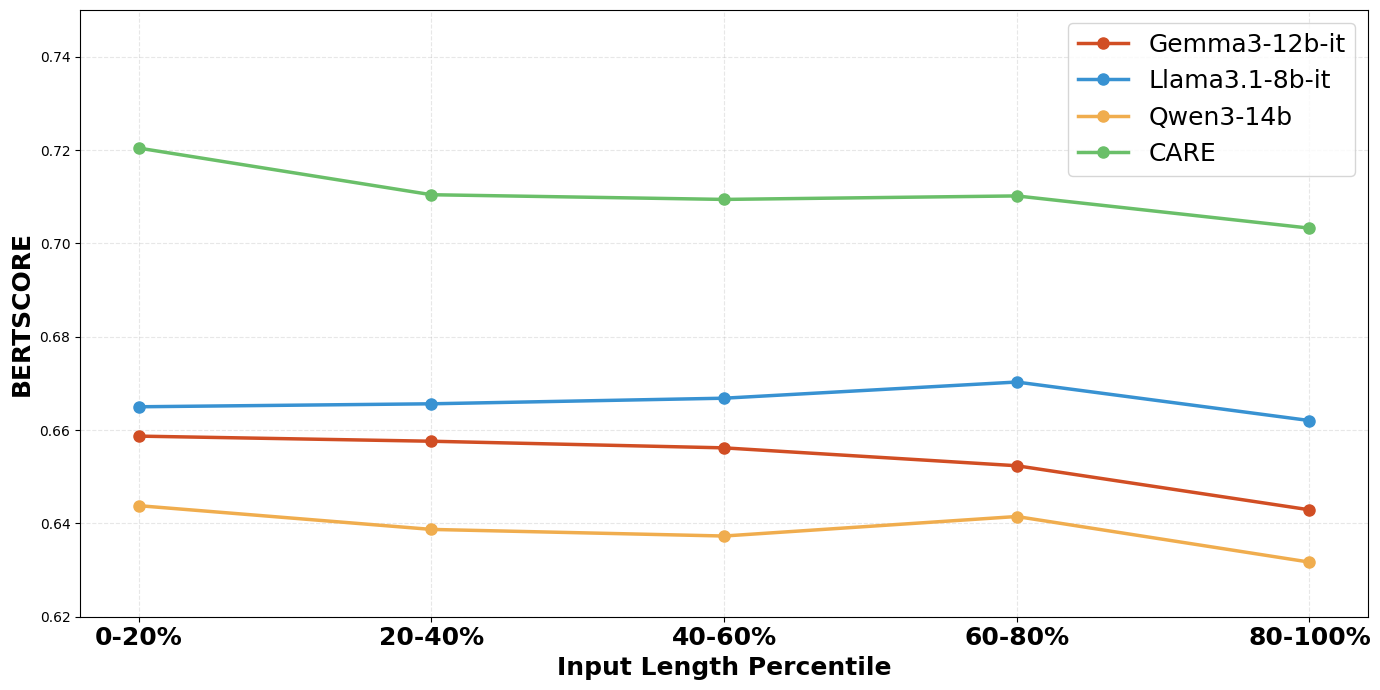}
    \caption{Comparison of Hebrew BERTScore performance across input length percentiles.} 
    \label{fig:BERTScore_sensitivity_analysis}
\end{figure}

We conducted a sensitivity analysis across input length percentiles to evaluate model robustness. To create these percentiles, the test set was sorted by input token length and divided into five equal quintiles (0--20\% to 80--100\%), where each bin represents 20\% of the data. As shown in Figure~\ref{fig:BERTScore_sensitivity_analysis}, CARE consistently outperforms all baselines in Hebrew, maintaining high semantic alignment (BERTScore) even as general-purpose models' performance degraded in longer contexts (80--100\%). 
The Arabic results exhibit a similar trend: CARE maintains superior clinical alignment compared to all baselines throughout the interaction.
These results  indicate the stability of the semantic alignment superiority of CARE over the baseline models across conversational history lengths. 

\section{Discussion}
Our experiments demonstrate that domain-specific fine-tuning using full conversation histories allows CARE to significantly outperform general-purpose LLMs in replicating the nuanced, strategic dialogue required for crisis intervention. These results suggest that the "cultural and tactical gap" observed in unadapted models is not merely a linguistic limitation, but a lack of alignment with specialized clinical protocols.

The performance of the CARE model suggests a distinct underlying mechanism. Providing the full conversation history enables the model to capture temporal dependencies and emotional trajectories, allowing it to generate responses that are both semantically and emotionally coherent. A particularly significant finding is the model's improvement in the Support Intent Match (SIM) metric. Despite the absence of explicit strategy labels during the training phase, CARE successfully increased significantly its SIM in both Hebrew and Arabic compared to baseline models. This indicates that CARE implicitly internalized professional counseling logic, learning when to utilize specific interventions like Reflection or Prompting based purely on the evolving state of the help-seeker.

The critical need for this domain-specific adaptation is best illustrated by the qualitative gap between model outputs. When presented with a seeker expressing profound distress, a vanilla LLM (Gemma3-12B-it in our case)  generated safe but detached clinical directives or analytical validation: 
\begin{itemize}
    \item \textit{``...I recommend that you reach out to a mental health professional. They can help you deal with your feelings...''}
    \item \textit{``I understand that you feel a lot of worry about the future... It shows a lot of self-awareness and a real desire to improve.''}
\end{itemize}

While factually sound, such responses often lack the emotional resonance required for immediate crisis de-escalation and can appear overly cognitive. In contrast, the CARE model produced responses that mirror the validating, reflective, and active tone of professional Sahar counselors: 
\begin{itemize}
    \item \textit{``Dear, I hear your disappointment.. maybe even your frustration.. I hear your desire to move on, to find other things that are indeed possible..''}
    \item \textit{``I hear how helpless you feel in the face of social anxiety...as if there is no way to deal with it...Can you think of one thing that will make you happy now?''}
\end{itemize}

This shift demonstrates how CARE's domain-specific training enables it to better capture elements of the emotional language and strategic flexibility exhibited by professional crisis counselors, producing more contextually appropriate crisis support responses.

\section{Limitations}
While CARE demonstrates a marked improvement over current baselines, several limitations remain. First, the Support Intent Match (SIM) metric is inherently dependent on the accuracy of the underlying Strategy Classifier. Although the classifier was fine-tuned on expert-annotated data, misclassifications in clinical intent could potentially introduce noise into the reported alignment scores.

Second, our evaluation relied primarily on automated metrics (ROUGE, BLEU, BERTScore, and PPL). While these serve as robust proxies for linguistic and semantic quality, they cannot fully substitute for real-time human expert evaluation of clinical safety and nuances in high-stakes crisis de-escalation. Finally, our dataset is specific to the Sahar organization’s methodology in Hebrew and Arabic. The results may vary in different cultural contexts or when applied to general psychotherapy sessions that do not follow the specific de-escalation protocols of a crisis hotline.

\section{Safety and Ethical Considerations}

The deployment of generative models in high-stakes crisis intervention requires a robust ethical framework to prioritize help-seeker safety and preserve counselor accountability. As recent research highlights, unaligned LLMs can inadvertently violate ethical standards in mental health practice through deceptive empathy or inadequate crisis management \cite{Iftikhar_Xiao_Ransom_Huang_Suresh_2025}. To address these risks, CARE is developed with the following ethical pillars:

\begin{itemize}
    \item \textbf{Human-in-the-Loop Decision Support:} The system is strictly designed as a decision-support tool rather than an autonomous agent. All generated suggestions are presented to trained counselors who retain full agency and responsibility for the final message sent to the help-seeker \cite{zalsman2021suicide}. This ensures that the intersubjective and relational dimensions of counseling are preserved.
    
    \item \textbf{Privacy and Anonymization:} To protect vulnerable populations, the training corpus underwent a rigorous two-tier anonymization process. In Hebrew, we utilized specialized de-identification tools for clinical text, while Arabic dialogues were processed using a fine-tuned NER model \cite{seker2021alephbertahebrewlargepretrained,antoun2020arabert}. These methods were supplemented by secondary regex-based filtering to ensure a robust privacy-preserving baseline.
    
    \item \textbf{Evaluation and Clinical Limitations:} While we demonstrate significant improvements across linguistic metrics such as ROUGE \cite{lin2004rouge}, BLEU \cite{papineni2002bleu}, and BERTScore\cite{zhang2020bertscoreevaluatingtextgeneration}, we acknowledge that text similarity does not substitute for clinical effectiveness. Future work must incorporate human-assessed generative tasks to fully validate the therapeutic alliance and clinical safety of the suggestions \cite{hua2024applyingevaluatinglargelanguage}.
\end{itemize}

\section{Conclusion}
This work presented CARE, a domain-adapted framework designed to align Large Language Models with the specialized strategies and supportive language of professional suicide-prevention counselors. By fine-tuning the Gemma-3-12b-it model on a curated corpus of high-effectiveness dialogues, we demonstrated that a training methodology leveraging full conversation histories significantly outperforms unadapted baselines in semantic fidelity, stylistic conformity, and strategic intent alignment.

Our results provide robust statistical evidence for the efficacy of domain-specific adaptation. CARE achieved a nearly ninefold increase in ROUGE-2 for Hebrew and a BLEU score in Arabic that is over 20 times higher than baseline averages. 
Furthermore, the model showed meaningful improvement in matching professional counseling strategies, with Support Intent Match (SIM) scores of 0.53 in Hebrew and 0.74 in Arabic. 
Paired Wilcoxon signed-rank tests with Holm correction confirm that these performance advantages are highly significant across all evaluation dimensions, effectively improving the cultural and tactical gap inherent in general-purpose models.

In conclusion, our study demonstrates that context-rich fine-tuning enables LLMs to produce recommendations that improve the empathetic style and supportive strategies essential to crisis intervention. Such systems offer a promising avenue for reducing response times, supporting overwhelmed counselors, and improving the accessibility of mental health support globally. Future work should explore hybrid approaches that integrate retrieval-augmented generation (RAG) for grounding responses in organizational guidelines as well as reinforcement learning with human feedback (RLHF) to further refine counselor alignment. Expanding evaluation to include real-time expert ratings from practicing counselors will be critical for validating real-world applicability.




\bibliographystyle{ACM-Reference-Format}
\bibliography{references}

\appendix

\end{document}